\documentclass{bmvc2k}

\usepackage{amsmath}
\usepackage{amssymb}
\usepackage{bbm}
\usepackage{tikz}
\usepackage{enumitem}
\usepackage{color}
\usepackage{comment}

\title{Crafting a multi-task CNN \\ for viewpoint estimation}

\addauthor{Francisco Massa}{http://imagine.enpc.fr/~suzano-f/}{1}
\addauthor{Renaud Marlet}{http://imagine.enpc.fr/~marletr/}{1}
\addauthor{Mathieu Aubry}{http://imagine.enpc.fr/~aubrym/}{1}

\addinstitution{
LIGM, UMR 814, Imagine,\\
Ecole des Ponts ParisTech, UPEM,ESIEE Paris, CNRS, UPE\\
Champs-sur-Marne, France
}

\runninghead{F. Massa et al.}{Crafting a multi-task CNN for viewpoint estimation}

\def\etal{\emph{et al}\bmvaOneDot}

\DeclareMathOperator*{\argmax}{arg\,max}

\newcommand{\back}[0]{{\mathrm{bg}}}
\newcommand{\obj}[0]{{\mathrm{obj}}}
\newcommand{\tdet}[0]{{\mathrm{det}}}
\newcommand{\vp}[0]{\mathrm{pose}}
\newcommand{\vect}[1]{\mathbf{{#1}}}

\newcommand{\indicator}[1]{\mathbbm{1}_{#1}}
\newcommand{\jreg}{{\textrm{j-reg}}}
\newcommand{\jclassif}{{\textrm{j-classif}}}
\newcommand{\tablefontsize}{\small}

\begin{document}

\maketitle

\begin{abstract}
 Convolutional Neural Networks (CNNs) were recently shown to provide state-of-the-art results for object category viewpoint estimation. However different ways of formulating this problem have been proposed and the competing approaches have been explored with very different design choices. This paper presents a comparison of these approaches in a unified setting as well as a detailed analysis of the key
  factors that impact performance. Followingly, we present a new joint training method with the detection task and demonstrate its benefit.
  We also highlight the superiority of classification approaches over regression approaches, quantify the benefits of deeper architectures and extended training data, and demonstrate that synthetic data is beneficial even when using ImageNet training data. 
  By combining all these elements, we demonstrate an improvement of approximately $5\%$ mAVP over previous state-of-the-art results on the Pascal3D+ dataset \cite{xiang2014beyond}. In particular for their most challenging 24 view classification task we improve the results from $31.1\%$ to $36.1\%$ mAVP. 
\end{abstract}

\section{Introduction}
Joint object detection and viewpoint estimation is a long-standing problem in computer vision. While it was initially tackled for single objects with known 3D models \cite{Roberts65,Lowe87,Huttenlocher87}, it was progressively investigated for complete object categories. The interest in this problem has recently increased both by the availability of the Pascal3D+ dataset \cite{xiang2014beyond}, which provides a standard way to compare algorithms on diverse classes, and by the improved performance of object detection, which encouraged researchers to focus on extracting more complex information from the images than the position of objects.

Convolutional Neural Networks were recently applied successfully to this task of object category pose estimation \cite{su2015render,tulsiani2015viewpoints}, leading to large improvements of state-of-the-art results on the Pascal3D+ benchmark. However many elements play an important role in the quality of these results, which have not yet been fully analyzed. In particular, several approaches have been proposed, such as a regression approach with joint training for detection \cite{osadchy07,penedones11}, a direct viewpoint classification \cite{tulsiani2015viewpoints}, and a geometric structure aware fine-grained viewpoint classification \cite{su2015render}, where the authors modify the classification objective to take into account the uncertainty of the annotations and encode implicitly the topology of the pose space. These papers however differ in a number of other ways, such as the training data or the network architecture they use, making it difficult to compare performances. We explore systematically the essential design choices for a CNN-based approach to pose estimation and we demonstrate that a number of elements influence the performance of the final algorithm in an important way.


\subsection*{Contributions} In this paper, we study several factors that affect performance for the task of joint object detection and pose estimation with CNNs. Using the best design options, we rationally define an effective method to integrate detection and viewpoint estimation, quantify its benefits, as well as the boost given by deeper networks and more training data, including data from ImageNet and synthetic data. 
We demonstrate that the combination of all these elements leads to an important improvement over state-of-the-art results on Pascal3D+, from $31.1\%$ to $36.1\%$ AVP in the case of the most challenging 24 viewpoints classification. While several of the elements that we employ have been used in previous work \cite{penedones11,su2015render,tulsiani2015viewpoints}, we know of no systematic study of their respective and combined effect, resulting in an absence of clear good practices for viewpoint estimation and sub-optimal performances.
Our code is available at \url{http://imagine.enpc.fr/~suzano-f/bmvc2016-pose/}.

\subsection*{Related work}
\label{subsec:related_work}

\paragraph{Convolutional Neural Networks.}
While convolutional neural network have a long history in computer vision (e.g. \cite{lecun1989backpropagation}), their use has been generalized only in 2012 after the demonstration of their benefits by Krizhevsky \etal \cite{krizhevsky2012imagenet} on the ImageNet large-scale visual recognition challenge \cite{deng2009imagenet}. Since then, they have been used to increase performances on many vision tasks.



This has been true in particular for object detection, where the R-CNN technique of Girshick \etal \cite{girshick14CVPR} provided an important improvement over previous methods on the Pascal VOC dataset \cite{everingham2010pascal}. Relying on an independent method to provide bounding box proposals for the objects in the image, R-CNN fine-tunes a network pre-trained on ImageNet to classify these proposal as objects or background. This method has then been improved in several ways, in particular using better network architectures \cite{he2015deep}, better bounding box proposals \cite{NIPS2015_5852} and a better sharing of the computations inside an image \cite{kaiming14ECCV,girshick2015fast}. 

\paragraph{Viewpoint estimation.}
Rigid object viewpoint estimation was first tackled in the case of object instances with known 3D models, together with their detection \cite{Roberts65,Lowe87,Huttenlocher87,Arandjelovic11,Li12,lim2013parsing}. These approaches were extended to object categories detection using either extensions of Deformable Part Models (DPM) \cite{Felzenszwalb10,Glasner11,Hejrati12,Pepik12}, parametric models \cite{Zia13,Xiang12} or large 3D instances collections \cite{aubry2014seeing,su2015render}.


With the advent of Pascal3D+ dataset \cite{xiang2014beyond}, which extends Pascal VOC dataset \cite{everingham2010pascal} by aligning a set of 3D CAD models for 12 rigid object classes, learning-based approaches using only on example images became possible and proved their superior performance. For example, Xiang \etal\cite{xiang2014beyond} extended the method of \cite{pepik2012teaching}, which uses an adaptation of DPM with 3D constraints to estimate the pose. CNN-based approaches, which were until the availability of the Pascal3D+ data limited to special cases such as faces \cite{osadchy07} and small datasets \cite{penedones11}, also began to be applied to this problem at a larger scale. In \cite{massa2014convolutional}, we explored different pose representations and showed the interest of joint training using AlexNet \cite{krizhevsky2012imagenet} and Pascal VOC \cite{everingham2010pascal} data. \cite{tulsiani2015viewpoints} used a simple classification approach with the VGG16 network \cite{simonyan2014very} and annotations for ImageNet objects and established the current state-of-the-art on Pascal3D+. \cite{su2015render} introduced a discrete but fine-grained formulation of the pose estimation which takes into account the geometry of the pose space, and demonstrate using AlexNet that adding rendered CAD models could improve the results over using Pascal VOC data alone.




\section{Overview}
\label{subsec:overview}
\vspace{-0.5em}
We focus on the problem of detecting and estimating the pose of objects in images, as defined by the Pascal3D+ challenge Average Viewpoint Precision (AVP) metric. In particular, we focus on the estimation of the azimuthal angle. For object detection, we use the standard Fast R-CNN framework \cite{girshick2015fast}, which relies on region proposal but is significantly faster than the original R-CNN \cite{girshick14CVPR}. 
In addition, we associate a viewpoint to each bounding box and for each object class. Indeed, since viewpoint conventions may not be coherent for the different classes, we learn a different estimator for each class. However, to avoid having to learn one network per class, we share all but the last layer of the network between the different classes.

In Section~\ref{sec:methods}, we first discuss different approaches to viewpoint prediction with CNNs and in particular the differences between regression and classification approaches. Then in Section~\ref{sec:joint}, we introduce different ways to integrate the viewpoint estimation and the detection problem. Finally, in Section~\ref{sec:exp} we present the results of the different methods as well as a detailed analysis of different factors that impact performance.

\paragraph{Notations.}
\vspace{-0.5em}
We call $N_s$ be the number of training samples and $N_c$ the number of object classes. 
For $i\in\left\lbrace 1, ... , N_s\right\rbrace$ we associate to the $i$-th training sample $x^i$ its azimuthal angle $\theta^i\in[0,2\pi[$, its class $c^i\in\left\lbrace 1, ... , N_c\right\rbrace$ and the output of the network with parameters $\vect{w}$, $f^{\vect{w}}(x^i)$. The viewpoints are often discretized and we call $N_v$ the number of bins, and
$\tilde\theta^i \in \left\lbrace 1, ... , N_v\right\rbrace$ the bin that includes $\theta^i$.
We use subscripts to denote the elements of a tensor; for example, $f^{\vect{w}}(x^i)_{k,l}$ is the element $(k,l)$ from tensor $f^{\vect{w}}(x^i)$.

\section{Approaches for viewpoint estimation}
\label{sec:methods}
\vspace{-0.5em}

\begin{figure}[t]
\centering
\subfigure[2D regression]{\includegraphics[width=0.2\linewidth]{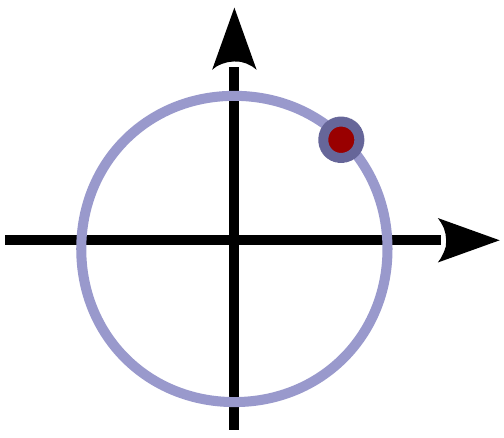}}
\subfigure[3D regression]{\includegraphics[width=0.2\linewidth]{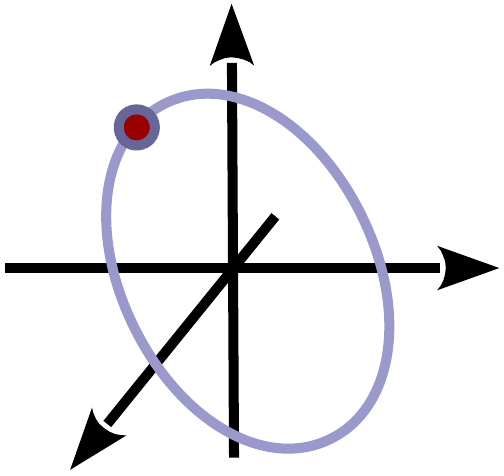}}
\subfigure[Direct classification]{
        \centering\includegraphics[width=0.22\linewidth]{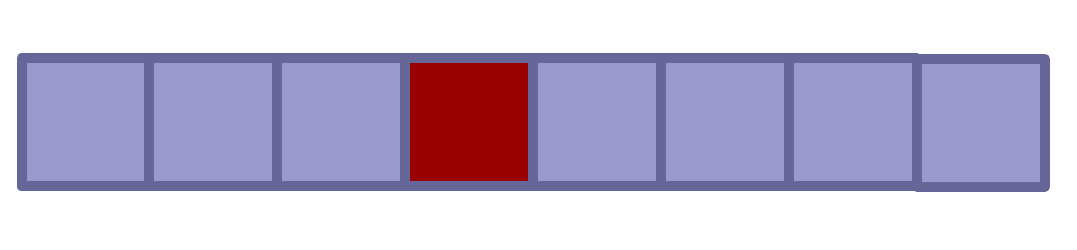}}
\subfigure[Geom. structure aware classification]{\includegraphics[width=0.18\linewidth]{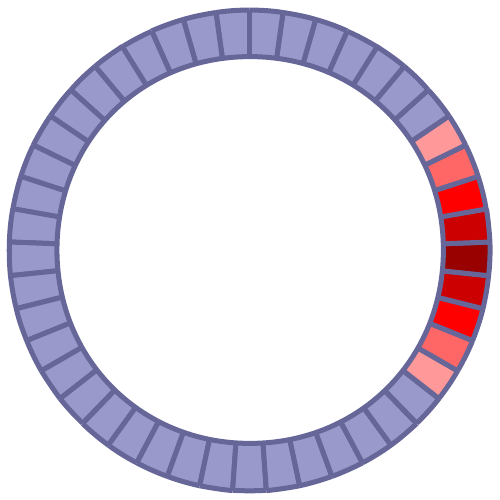}}
\caption{ Different approaches to orientation prediction discussed in this paper. The target for each approach is visualized in red. For the regression approaches, the possible values of the targets lie on a line. For the classification approaches, the predictions correspond to probability distributions on a discrete set.
}
\label{fig:app}
\vspace{-0.5em}
\end{figure}

In this section, we assume the bounding box and the class of the objects are known and we focus on the different approaches to estimate their pose.
Section~\ref{subsec:reg} first discusses the design of regression approaches.
Section~\ref{subsec:discrete} then presents two variants of classification approaches. The intuition behind these different approaches are visualized on Figure~\ref{fig:app}.

\subsection{Viewpoint estimation as regression}
\label{subsec:reg}

The azimuth angle of a viewpoint being a continuous quantity, it is natural to tackle pose estimation as a regression problem. 
The choice of the pose representation $F(\theta)$ of an azimuthal angle~$\theta$ is of course crucial for the effectiveness of this regression. Indeed, if we simply consider $F(\theta)=\theta$, the periodicity of the pose is not taken into account.
Thus, as highlighted in \cite{osadchy07}, a good pose representation $F(\theta)$ satisfies the following properties: (a) it is invariant to the periodicity of the angle $\theta$, and (b) it is analytically invertible.

We explore two representations which satisfy both properties:
\begin{enumerate}[label=(\roman*)]
  \item
  $
  F(\theta) =
  \begin{bmatrix}
    \cos(\theta) ,\  \sin(\theta)
  \end{bmatrix}$, probably the simplest way to represent orientations, used for example in \cite{penedones11};
  \item
  $
  F(\theta) =
  \begin{bmatrix}
    \cos\left(\theta - \frac{\pi}{3}\right) ,\
    \cos\left(\theta\right) ,\
    \cos\left(\theta + \frac{\pi}{3}\right)
  \end{bmatrix}
  $, a formulation which was presented in \cite{osadchy07}, and that has a higher dimensionality than the previous one, allowing more flexibility for the network to better capture the pose information.
\end{enumerate}
These representations have different output dimensionality $N_{d}$, respectively 2 and 3, and we designate the associated regressions by {\it regression 2D} and {\it regression 3D} respectively. Since we treat the regression independently for each class, the outputs $f^\vect{w}(x)$ of the network that we train for pose estimation have values in $\mathbb{R}^{N_c\times N_{d}}$ and we designate by $f^\vect{w}(x)_{c,k}$ the angular element $k$ of the output for class $c$. 

For training the regression with these representations, we used the Huber loss (also known as Smooth L1) on each component of the pose representation $F(\theta)$. It is known to be more robust to outliers than the Euclidean loss and provides much better results in our experiments. Our regression loss can then be written:
\vspace{-0.5em}
\begin{equation}
\label{eq:reg_loss}
  L^{\mathrm{reg}}(\vect{w}) = \sum_{i=1}^{N_s}\sum_{k=1}^{N_{d}}  H (f^\vect{w}(x^i)_{c^i,k} - F(\theta^i)_k)
  \vspace{-0.5em}
\end{equation}
with $H$ the Huber loss.
Given the output $f^\vect{w}(x)_{c,{\scriptscriptstyle\bullet}}$ of the network for a sample $x$ of class $c$, we can estimate its pose simply by computing the pose of the closest point on the curve described by $F$ (cf.\ Figure~\ref{fig:app}). Other regression approaches and loss are discussed in \cite{massa2014convolutional} but lead to lower performances.

\subsection{Viewpoint estimation as classification}
\label{subsec:discrete}
As pointed out by \cite{su2015render}, the main limitation of a regression approach to viewpoint estimation is that it cannot represent well the ambiguities that may exist between different viewpoints. Indeed, objects such as a table have symmetries or near symmetries that make the viewpoint estimation problem intrinsically ambiguous, and this ambiguity is not well handled by the representations discussed in the previous paragraph. One solution to this problem is to discretize the pose space and predict a probability for each orientation bin, thus formulating the problem as one of classification. Note that a similar difficulty is found in the problem of keypoint prediction, for which the similar solution of predicting a heat map for each keypoint instead of predicting directly its position has proven successful \cite{tulsiani2015viewpoints}.



In the case of a classification approach, the output of the network belongs to $\mathbb{R}^{N_c\times N_v}$ and each value can be interpreted as a log-probability. We write $f^{\vect{w}}(x)_{c,v}$ the value corresponding to the orientation bin $v$ for an input $x$ of class $c$.
\subsubsection{Direct classification}
\label{subsec:direct}

The approach successfully applied in \cite{tulsiani2015viewpoints} is to simply predict, for each class independently, the bin in which the orientation of the object falls. This classification problem can be addressed for each object class with the standard cross-entropy loss:
\begin{equation}
  \label{eq:cross_entropy}
  L^{\mathrm{classif}}(\vect{w}) = -  \sum_{i=1}^{N_s}   \log\left(\frac{\exp(f^{\vect{w}}(x^i)_{c^i,\tilde{\theta}^i})}{\sum_{v=1}^{N_v} \exp(f^{\vect{w}}(x^i)_{c^i,v})}\right)
  \vspace{-0.5em}
\end{equation}
%
%
%
%
At test time,
the predicted angular bin $\hat{\theta}(x,c)$ for an input $x$ of class $c$ is given by
\begin{equation}
\label{eq:inf_class}
  \hat{\theta}(x,c) = \argmax_{v\in\{1,\ldots,N_v\}} f^{\vect{w}}(x)_{c,v} 
  \vspace{-0.5em}
\end{equation}

\subsubsection{Geometric structure aware classification}
\label{subsec:rendercnn}
The drawback of the previous classification approach is that it learns to predict the poses without using explicitly the continuity between close viewpoints. Two neighboring bins have indeed a lot in common. This geometrical information may be especially important for fine-grained orientation prediction, where only few examples per bin are available.  

A solution to this problem was proposed in \cite{su2015render}. The authors finely discretize the orientations in $N_v=360$ bins and consider the angle estimation as a classification problem, but adapt the loss to include a structured relation between neighboring bins and penalize less angle errors that are smaller:

\begin{equation}
\label{eq:geom}
  L^{\mathrm{geom}} (\vect{w}) =-  \sum_{i=1}^{N_s} \sum_{v=1}^{N_v}  \exp \left(\frac{- d(v,\tilde{\theta}^i)}{\sigma} \right)\log\left(\frac{\exp(f^{\vect{w}}(x^i)_{c^i,v})}{\sum_{v=1}^{N_v} \exp(f^{\vect{w}}(x^i)_{c^i,v})}\right)
\end{equation}
where $d(v,\tilde{\theta}^i)$ is the distance between the centers of the two bins $v$ and $\tilde{\theta}^i$, and $\sigma$ is a parameter controlling how much similarity is enforced between neighboring bins. Following \cite{su2015render}, we use $\sigma=3$ for $N_v=360$. The inference is done as in Equation~\eqref{eq:inf_class}.

\section{Joint detection and pose estimation}
\label{sec:joint}
The methods presented in the previous section assume that the object detector is already trained and kept independent from the pose estimator. Since object detection and pose estimation relies on related information, we expect a benefit from training them jointly. We thus present extensions of the methods from Section~\ref{sec:methods} to perform this joint training.

\subsection{Joint model with regression}
Two main approaches can be considered to extend the regression approach of Section~\ref{subsec:reg} to jointly perform detection. The first one, described in \cite{osadchy07} is to encode respectively the presence or absence of an object by a point close or far from the regression line described by $F$ in the space where the regression is performed. An alternative approach, discussed in \cite{penedones11}, is to add an output to the regression network specifically dedicated to detection. The loss used to train the network can then be decomposed into two terms: a classification loss $L^{\mathrm{\det}}(\vect{w})$, which is independent on the pose, and a regression loss $L^{\mathrm{reg}}(\vect{w})$ which takes into account only the pose estimation. Since state-of-the-art performance for detection are obtained using a classification loss, we selected the second option in the following.

Our network thus has two outputs: $ f^{\vect{w},\tdet} (x)\in \mathbb{R}^{N_c+1}$ for the detection part (predicting probabilities for each of the $N_c$ classes and the background class), and $f^{\vect{w},\vp} (x) \in \mathbb{R}^{N_c\times N_d}$ for the pose estimation part. 
The multi-task loss for joint classification and regression-based pose estimation writes as follows:
\begin{equation}
  \label{eq:joint_reg}
  L^\jreg(\vect{w}) = L^{\mathrm{det}}(\vect{w}) + \lambda L^{\mathrm{reg}}(\vect{w})
\end{equation}
We define $L^{\mathrm{reg}}$ exactly as in Equation~\eqref{eq:reg_loss}, using the pose estimation output of the network $f^{\vect{w},\vp} (x)$. The detection loss $L^{\mathrm{det}}$ is the standard cross-entropy loss for detection, using the detection part of the network output $ f^{\vect{w},\tdet} (x)$. 
We set the balancing parameter $\lambda=1$ in our experiments.



Also, we share the weights of the detection and pose estimation network only up to the \emph{pool5} layer.
This is essential to obtain a good performance, as the regression and classification losses are different enough that sharing more weights leads to much worse results.

\subsection{Joint model with classification}
\label{subsec:joint_discrete}
A similar approach, separating two branches of the network, can be applied for classification. However, we introduce a new simpler and parameter-free way to perform jointly detection and pose estimation in a classification setup. Indeed, one can simply add a component, associated to the background patches, to the output vector of the pose estimation setup of Section~\ref{subsec:discrete} and normalize it globally, rather than for each class independently as in Equation~\eqref{eq:cross_entropy}. Each value is then interpreted as a log probability of the object being of one class and in a given orientation bin, rather than the conditional probability of the object being in a given orientation bin knowing its class. To obtain the probability of the object to belong to one class, one can simply sum the probabilities corresponding to all the bins for this class.  

Similar to Section~\ref{subsec:discrete}, we write $f^{\vect{w},\obj}(x)_{c,v}\in \mathbb{R}^{N_c\times N_v}$ the value of the network output corresponding to the orientation bin $v$ for an input $x$ of class $c$. We additionally write $f^{\vect{w}, \back}(x)\in \mathbb{R}$ its value corresponding to the background and associate a class $c^i=0$ to the elements $x^i$ in the background.
The loss, which derives from the cross-entropy, writes:
\begin{multline}
  \label{eq:joint_discrete}
   L^{\jclassif}(\vect{w}) =
   -  \sum_{i=1}^{N_s}  \indicator{c^i=0}\log\left(\dfrac{\exp(f^{\vect{w},\back}(x^i))}{\exp(f^{\vect{w},\back}(x^i))+\sum_{c=1}^{N_c} \sum_{v=1}^{N_v} \exp(f^{\vect{w},\obj}(x^i)_{c,v})}\right)\\
   -  \sum_{i=1}^{N_s}  \indicator{c^i\neq0}\log\left(\dfrac{\exp(f^{\vect{w},\obj}(x^i)_{c^i,\tilde{\theta}^i})}{\exp(f^{\vect{w},\back}(x^i))+\sum_{c=1}^{N_c} \sum_{v=1}^{N_v} \exp(f^{\vect{w},\obj}(x^i)_{c,v})}\right)
\end{multline}
At inference, the score associated to the detection of an object $x$ for class $c$ is
\begin{equation}
S(x,c)=\dfrac{\sum_{v=1}^{N_v}\exp(f^{\vect{w},\obj}(x)_{c,v})}{\exp(f^{\vect{w},\back}(x))+\sum_{c'=1}^{N_c} \sum_{v=1}^{N_v} \exp(f^{\vect{w},\obj}(x)_{c',v})}
\end{equation}


\section{Experiments}
\label{sec:exp}
We now present experiments comparing the different approaches for pose estimation which were presented in the previous sections.
Our experiments are based on the Fast~R-CNN object detection framework \cite{girshick2015fast}
, with Deep Mask \cite{NIPS2015_5852} bounding boxes proposals.


We trained and evaluated our models using the Pascal3D+ dataset \cite{xiang2014beyond}, which contains pose annotations for the training and validation images from Pascal VOC 2012 \cite{everingham2010pascal} for 12 rigid classes, as well as for a subset of ImageNet \cite{deng2009imagenet}. We also extended the training data by adding the synthetic images from \cite{su2015render}.
The evaluation metric we used is the \emph{Average Viewpoint Precision} (AVP) associated to Pascal3D, which is very similar to the standard Average Precision (AP) metric used in detection tasks, but which considers as positive only the detections for which the viewpoint estimate is correct. More precisely, the viewpoints are discretized into $K$ bins and the viewpoint estimate is considered correct if it falls in the same bin as the ground-truth annotation. We focus on the AVP24 metric, 
which discretizes the orientation into $K=24$ bins and is the most fine-grained of the Pascal3D+ challenge \cite{xiang2014beyond}. 
We also consider the mean AP (mAP) and mean AVP (mAVP) over all classes.

\subsection{Training details}
We fine-tuned our networks, starting from a network trained for ImageNet classification, using Stochastic Gradient Descent with a momentum of 0.9 and a weight decay of 0.0005. We augment all datasets with the horizontally-flipped versions of each image, flipping the target orientations accordingly. During the training of the joint detection and pose estimation models, 25\% of the mini-batches consist of positive examples. Our mini-batches are of size 128 except when using synthetic images. When using synthetic images, we randomly create montages with the rendered views from \cite{su2015render}, each montage containing 9 objects, for a total mini-batch size of 137 (96 backgrounds and 32 positive patches from real images and 9 positive synthetic objects). This allows for an efficient training in the setup of Fast R-CNN.

We initialized the learning rate at 0.001, and divided it by 10 after convergence of the training error. The number of iterations depends of the amount of training data: when using only Pascal VOC data, we decrease the learning rate after 30K iterations and continue to train until 40K; when adding ImageNet data  we decrease the learning rate after 45K iterations and continue to train until 100K; and finally, when adding synthetic data, we decrease the learning rate after 100K iterations and continue to train until 300K. 

All experiments were conducted using the Torch7 framework \cite{torch7_NIPSworkshop2011} and we will release our full code upon publication.

\begin{table}
  \centering
  \tablefontsize
  \caption{Different approaches for pose estimation with AlexNet architecture, Pascal VOC 2012 data, and using a fixed detector}
  \label{tab:methods_noclassif}
  \begin{tabular}{c | c c}
    Method & mAP & mAVP\@24\\
    \hline
    Regression 2D & 51.6 & 13.9 \\
    Regression 3D & 51.6 & 15.7 \\
    Direct classification & 51.6 & {\bf 19.3} \\
    Geometric structure aware classification  & 51.6 & 18.4
  \end{tabular}
\end{table}

\subsection{Results}

\subsubsection{Comparison of the different approaches for pose estimation}
We first compare the different approaches for pose estimation from Section~\ref{sec:methods}. We use a fixed object detector based on the AlexNet architecture, trained for detection on Pascal VOC 2012 training set and we report the results in Table~\ref{tab:methods_noclassif}.
We can first observe that for regression, a pose representation with a higher dimensionality (3D) performs better than when using a smaller dimensionality (2D). We believe 
the redundancy in the representation helps to better handle ambiguities in the estimation. 
The classification approach however significantly outperforms both regressions ($19.3\%$ AVP compared to $13.9\%$ and $15.7\%$). 
Interestingly, the simplest classification approach from Section~\ref{subsec:discrete} performs slightly better than the geometry-aware method. We think the main reason for this difference is that the simple classification optimize exactly for the objective evaluated by the AVP, and thus this result can be seen as an artefact of the evaluation. Note that the results could be different for even more fine-grained estimation where less examples per class are available. Nevertheless, since the more complex geometric structure aware approach performed worse than the direct classification baseline, we focus in the rest of this paper on the simplest direct classification approach.



\begin{table}
  \centering
  \tablefontsize
  \caption{Jointly training for detection and pose estimation with AlexNet architecture and Pascal VOC 2012 data}
  \label{tab:methods_joint}
  \begin{tabular}{c | c c | c c}
    & \multicolumn{2}{c|}{Joint detector} & \multicolumn{2}{c}{Independant detector} \\
    Method & mAP & mAVP\@24 & mAP & mAVP\@24\\
    \hline
    Joint Regression 2D & 49.2 & 15.7 & 51.6 & 16.4\\
    Joint Regression 3D & 49.6 & 17.1 & 51.6 & 17.4 \\ 
    Joint classification & 48.6 & {\bf 21.1} & 51.6 & {\bf 20.5} \\
  \end{tabular}
  \vspace{-1em}
\end{table}

\subsubsection{Benefits of joint training for detection and pose estimation}
We evaluate the benefits of jointly training a model to detect the objects and predict their orientation. These benefits can be of two kinds. First, the order of the detections candidates given by the new detector may favor the confident orientations and thus increase the AVP. Second, the pose estimates can be better for a given object. To evaluate both effects independently, we report in Table~\ref{tab:methods_joint} the results using both the order given by the detector used in the previous section and the order given by the new joint classifier. All experiments were performed as above, 
with the AlexNet architecture and the Pascal VOC training data.

Comparing Table~\ref{tab:methods_joint} to Table~\ref{tab:methods_noclassif} shows two main effects. First, the mAVP is improved even when using the same classifier, demonstrating improved viewpoint estimation with joint training. Second, the mAP is decreased, showing that the detection performs worse when trained jointly. However, one can also notice that the best mAVP is still obtained with the joint classifier. This shows that the pose estimation is better in the joint model, and also that for the case of classification the order learned when training jointly the detector favours confident poses. This is not the case for the regression approaches for which the best results are obtained using the independent detector and the jointly-learned pose estimation.

\subsubsection{Influence of network architectures and training data}

\begin{table}
  \vspace{0.5em}
  \centering
  \tablefontsize
  \caption{Influence of the amount of training data and network architecture on our joint classification approach}
  \label{tab:dataset_size}
  \begin{tabular}{c | c c | c c}
     & \multicolumn{2}{c|}{AlexNet} & \multicolumn{2}{c}{VGG16} \\
    Training data & mAP & mAVP24 & mAP & mAVP24\\
    \hline
    Pascal VOC2012 train &48.6 &21.1 & 56.9 & 27.3 \\
    + 250 per class &51.6 &25.0 & 58.0 & 30.0 \\
    + 500 per class &53.8 &26.5 & 59.0 & 31.6 \\
    + 1000 per class &53.6 &28.3 & 60.0 & 32.9 \\
    + full ImageNet &52.8 & 28.4 & 59.9 & 34.4 \\
    + synthetic data & 55.9 & {\bf 31.5} &  61.6 & {\bf 36.1} \\
  \end{tabular}
  \vspace{-1em}
\end{table}

In this section, we consider our joint classification approach, which performs best in the evaluations of the previous section, and study how its performance varies when using different architectures and more training data. 

The comparison of the left and right columns of Table~\ref{tab:dataset_size} shows that unsurprisingly the use of the VGG16 network instead of AlexNet consistently improves performances. This improvement is slightly less for the mAVP than for the mAP, hinting that the mAVP boost is mainly due to improved detection performances.

For the training data, we first progressively add training images from ImageNet to the training images from Pascal VOC. The full subset of the ImageNet dataset annotated in Pascal3D+ contains in average approximately 1900 more images per class, but is strongly unbalanced between the different classes. The analysis of these results shows consistent improvements when the training set includes more data. Interestingly, the mAVP is improved more than the mAP, showing that the additional data is more useful for pose estimation than for detection. The addition of synthetic data (2.4M positive examples) improves the results even more, demonstrating that the amount of training data is still a limiting factor even if one uses an AlexNet architecture and includes the ImageNet images, a fact that was not demonstrated in \cite{su2015render}. Note that our joint approach significantly outperforms the state-of-the-art results \cite{tulsiani2015viewpoints} (currently $31.1\%$ mAVP, based on VGG16 and ImageNet annotations) both without using synthetic data with VGG16, and with synthetic data and AlexNet architecture.

\subsubsection{Comparison to the state of the art}
\label{subsubsec:summary}
Table~\ref{tab:summary} provides the details of the AVP24 performance improvements over all classes as well as a comparison with three baselines: DPM-VOC+VP \cite{pepik2012teaching}, which uses a modified version of DPM to also predict poses, Render for CNN \cite{su2015render} which uses real images from Pascal VOC as well as CAD renders for training a CNN based on AlexNet, and \cite{tulsiani2015viewpoints} which uses a VGG16 architecture and ImageNet data to classify orientations for each object category. It can be seen that we improve consistently on all baselines except for the chair class. A more detailed analysis shows that this exception is related to the difference between the ImageNet and Pascal chairs. Indeed, when adding the ImageNet data to the Pascal data, the detection performance for chairs drops from $34.5\%$ AP to $19.23\%$ AP. Similarly, the difference between the very different appearance of the rendered 3D models and real images is responsible for the fact that synthetic training data decreases performance on boats, motorbikes and trains. In average, we still found that synthetic images boost the results by $1.7\%$ mAVP.  \\

Finaly, Table~\ref{tab:summary2} provides the comparison between our full pipeline and the baselines for the 4, 8 and 16 viewpoint classification tasks, showing that our improvement of the state of the art is consistently high.

\section{Conclusion}
Combining our joint classification approach to the improvements provided by a deep architecture and additional training data, we increase state-of-the-art performance of pose estimation by $5\%$ mAVP. We think that highlighting the different factors of this improvement and setting a new baseline will help and stimulate further work on viewpoint estimation.


\begin{table*}[t!]
  \caption{Summary of results and comparison with baselines using AVP24}
  \label{tab:summary}
  \centering
  \setlength{\tabcolsep}{2.3pt}{
  \footnotesize
  \vspace{0.5em}
  \begin{tabular}{c|c c c c c c c c c c c | c}
    \hline
  Method  & aero & bike & boat & bus & car & chair & table & mbike & sofa & train & tv & mAVP24 \\
  \hline
  DPM-VOC+VP \cite{pepik2012teaching} & 9.7 & 16.7 & 2.2 & 42.1 & 24.6 & 4.2 & 2.1 & 10.5 & 4.1 & 20.7 & 12.9 & 13.6 \\
  Render For CNN \cite{su2015render} & 21.5 & 22.0 & 4.1 & 38.6 & 25.5 & 7.4 & 11.0 & 24.4 & 15.0 & 28.0 & 19.8 & 19.8 \\
  Viewpoints \& Keypoints \cite{tulsiani2015viewpoints} & 37.0 &	33.4 &	10.0 & 	54.1 &	40.0 &	{\bf 17.5} &	19.9 &	34.3 &	28.9 &	43.9 &	22.7 &	31.1 \\
  \hline
  Classif. approach \& AlexNet & 21.6 & 15.4 & 5.6 & 41.2 & 26.4 & 7.3 & 9.3 & 15.3 & 13.5 & 32.9 & 24.3 & 19.3 \\
  + our joint training
  & 24.4 & 16.2 & 4.7 & 49.2 & 25.1 & 7.7 & 10.3 & 17.7 & 14.8 & 36.6 & 25.6 & 21.1 \\
  + VGG16 instead of AlexNet & 26.3 & 29.0 & 8.2 & 56.4 & 36.3 & 13.9 & 14.9 & 27.7 & 20.2 & 41.5 & 26.2 & 27.3 \\
  + ImageNet data & 42.4 & 37.0 & {\bf 18.0} & 59.6 & 43.3 & 7.6 & 25.1 & {\bf 39.3} & 29.4 & {\bf 48.1} & 28.4 & 34.4 \\
  + synthetic data & {\bf 43.2} & {\bf 39.4} & 16.8 & {\bf 61.0 }& {\bf 44.2} & 13.5 & {\bf 29.4} & 37.5 & {\bf 33.5} & 46.6 & {\bf 32.5} & {\bf 36.1} \\
  \end{tabular}
  }
\end{table*}

\begin{table*}[t!]
  \caption{Comparison with state of the art using AVP4, AVP8 and AVP16}
  \label{tab:summary2}
  \centering
  \setlength{\tabcolsep}{2.3pt}{
  \footnotesize
  \vspace{0.5em}
  \begin{tabular}{c|c | c c c c c c c c c c c | c}
    \hline
  Method  & measure & aero & bike & boat & bus & car & chair & table & mbike & sofa & train & tv & mAVP\\
  \hline
   \cite{tulsiani2015viewpoints} & AVP4 &63.1&	59.4&	23.0& 69.8&	55.2&	{\bf25.1}&	24.3	&61.1&	43.8	&59.4&	55.4 &49.1\\
  Ours &  AVP4 & {\bf 70.3        } & {\bf 67.0      } & {\bf 36.7    } & {\bf 75.4    } & {\bf 58.3    } & { 21.4    } & {\bf 34.5          } & {\bf 71.5        } & {\bf 46.0    } & {\bf 64.3    } & {\bf 63.4    } & {\bf  55.4}\\
  \hline
   \cite{tulsiani2015viewpoints} & AVP8 & 57.5	&54.8&	18.9	&	59.4	&51.5&{\bf 	24.7}&	20.5&	59.5	&43.7	&53.3&	45.6	&44.5\\
  Ours &  AVP8 &{\bf  66.0        } & {\bf 62.5      } & {\bf 31.2    } & {\bf 68.7    } & {\bf 55.7    } & { 19.2    } & {\bf 31.9          } & {\bf 64.0        } & {\bf 44.7    } & {\bf 61.8    } & {\bf 58.0    } & {\bf 51.3 }\\
  \hline

   \cite{tulsiani2015viewpoints} & AVP16 & 46.6&	42.0	&12.7	&	64.6	&42.7&{\bf 	20.8}	&18.5	&38.8	&33.5&	42.5&	32.9&	36.0 \\
  Ours &  AVP16 & {\bf 51.4       } & {\bf 43.0      } & {\bf 23.6      } & {\bf 68.9    } & {\bf 46.3    } & { 15.2    } & {\bf 29.3      } & {\bf 49.4        } & {\bf 35.6    } & {\bf 47.0    } & {\bf 37.3    } & {\bf 40.6 } \\

  \end{tabular}
  }
\vspace{-1em}
\end{table*}

\bibliography{refs}
\end{document}